\documentclass[conference]{IEEEtran}
\IEEEoverridecommandlockouts
\usepackage{cite}
\usepackage{amsmath,amssymb,amsfonts}
\usepackage{algorithmic}
\usepackage{graphicx}
\usepackage{textcomp}
\usepackage{xcolor}

\usepackage{makecell} 
\usepackage{listings}
\def\BibTeX{{\rm B\kern-.05em{\sc i\kern-.025em b}\kern-.08em
    T\kern-.1667em\lower.7ex\hbox{E}\kern-.125emX}}

\usepackage{fancyhdr}
\fancypagestyle{firstpage}{
  \fancyhf{}
  
  \fancyfoot[C]{\scriptsize
    \copyright~2026 IEEE. Personal use of this material is permitted. Permission from IEEE must be obtained for all other uses, in any current or future media, including reprinting/republishing this material for advertising or promotional purposes, creating new collective works, for resale or redistribution to servers or lists, or reuse of any copyrighted component of this work in other works.\\
    Published in \textit{2026 IEEE 50th Annual Computers, Software, and Applications Conference (COMPSAC)}, DOI: 10.1109/COMPSAC69091.2026.00315
  }
}

\begin{document}

\title{Towards Proactive Air Traffic Safety with Speech LLMs: Transcription, Attribute Tagging, and Readback Detection}




\author{
\IEEEauthorblockN{Raymond Chung}
\IEEEauthorblockA{
\textit{Logistics and Supply Chain MultiTech R\&D Centre} \\
Hong Kong, China \\
rchung@lscm.hk}
}

\maketitle
\thispagestyle{firstpage}
\enlargethispage{-1.2cm}

\begin{abstract}

Air traffic communication (ATC) is a highly specialized domain where noisy acoustic conditions, rapid speech rates, and domain‑specific terminology pose significant challenges for automatic speech recognition. We investigate the use of Speech Large Language Models (Speech LLMs) to address these challenges by combining accurate transcription with structured information extraction. We leverage publicly available ATC datasets and apply data augmentation together with metadata‑enriched training, which improves multi‑turn dialogue handling and reduces transcription errors. On the test set, our model achieved a word error rate of 16.70\% and a role classification accuracy of 96.60\%. Furthermore, we propose a system for extracting callsigns, commands, and values, enabling automated readback detection. Case studies show that the system can determine whether pilots correctly repeat critical instructions, thereby supporting safety monitoring. Taken together, these results demonstrate the potential of Speech LLMs to deliver reproducible transcription accuracy, reliable role attribution, and proactive safety monitoring in air traffic communication.

\end{abstract}


\section{Introduction}
\IEEEpubidadjcol
ATC is a safety‑critical domain where clarity and accuracy directly affect both operational efficiency and flight safety. Unlike casual dialogue, ATC exchanges follow highly structured phraseology, yet they are transmitted over noisy radio channels that complicate automatic transcription. Additional challenges arise from diverse speaker accents and rapid delivery rates, creating a demanding environment for automatic speech recognition (ASR) systems.

Every ATC dialogue involves at least two participants: the air traffic controller (ATCO) and the pilot. Distinguishing between these roles is essential because the operational meaning of an utterance depends on who is speaking. A clearance issued by an ATCO must be interpreted as an instruction, while the same phrase spoken by a pilot typically represents a readback or confirmation.

To address these challenges, a system must not only transcribe reliably but also produce structured outputs that capture speaker roles and operational attributes. Such processing enables systematic comparison between controller and pilot utterances and provides a foundation for downstream applications including automated readback detection, compliance monitoring, and incident analysis, thereby strengthening both safety and efficiency in ATC communication.

In this paper, we investigate the application of Speech LLMs to ATC communication, focusing on transcription, role identification, and attribute extraction. Leveraging on the domain‑specific datasets, we design methods to enhance robustness through data augmentation and metadata‑enriched training. By unifying transcription and structured tagging within a single framework, our approach enables automated readback detection and illustrates the potential of Speech LLMs to strengthen operational safety and reliability in air traffic control.

Our work makes the following contributions:
\begin{itemize}

    \item \textbf{Domain‑specific dataset and baseline comparison:} Using publicly available ATC audio datasets, we conducted a reproducible comparative evaluation of transcription error rates between Transformer‑based ASR and Speech LLM architectures. In particular, we evaluated the filler ``ah'' and found that its transcription is not required for ATC‑domain tasks, since omitting it yields cleaner outputs and improved recognition accuracy.
    
    \item \textbf{Integrated modeling for transcription and attribute tagging:} We develop methods that combine Speech LLMs with data augmentation and metadata‑enriched training, enabling more reliable transcription and accurate role assignment in multi‑turn dialogues. Structured extraction of callsigns, commands, and values is incorporated to support downstream applications.
    \item \textbf{Framework for readback detection:} We introduce a system that transcribes ATC exchanges and extracts callsigns, commands, and values, enabling automated comparison between controller and pilot utterances. Case studies demonstrate that this framework can proactively support readback error detection, contributing to aviation safety monitoring.
\end{itemize}

Together, these contributions highlight how Speech LLMs advance transcription accuracy, role understanding, and proactive safety monitoring in complex ATC environments.

\section{Related Work}

Recent work on ATC speech recognition and readback detection has explored both pipeline‑based architectures and integrated modeling approaches.

Khalil et al.~\cite{khalil2023automatic} introduced an automated speaker clustering pipeline tailored to the ATC domain. Their approach combined Speech Activity Detection (SAD) to separate speech from non‑speech, an ASR engine for transcription, and a sequence labeling model for role assignment. For classification, they fine‑tuned a BERT‑based model~\cite{devlin2019bert}, where BERT (Bidirectional Encoder Representations from Transformers) provides contextualized embeddings for text. This achieved an F1 score of 0.87 on human transcripts but only 0.53 when relying on ASR outputs, underscoring how upstream transcription errors propagate through multi‑stage pipelines and degrade downstream role assignment.

Blatt et al.~\cite{blatt2024joint} compared joint versus sequential architectures for Speaker Role Detection (SRD) and ASR in ATC communications. Leveraging pre‑trained self‑supervised encoders such as Wav2Vec 2.0~\cite{baevski2020wav2vec} and Serialized Output Training (SOT), their joint SRD‑ASR model consistently outperformed sequential pipelines, achieving lower word error rate (WER) and demonstrating effective few‑shot adaptation to new airport datasets. Although their experiments used relatively small public datasets (fewer than 2,000 utterances, with WERs above 20\%), the results highlight the effectiveness of joint modeling for role‑aware transcription.

Xu et al.~\cite{xu2025joint} extended the SOT paradigm by applying Whisper‑large‑v3~\cite{radford2023robust}, then the state of the art in ASR, for joint transcription and role tagging. Their experiments on adult–child conversations and the MMGSG dataset showed that encoder fine‑tuning improved both word error rate and role assignment. While we also prepend role markers to the transcription stream, our work advances this approach in the more demanding ATC domain by integrating role‑aware transcription with audio‑language models to address domain‑specific acoustic and linguistic variability.

Wee et al.~\cite{wee2025adapting} curated a Southeast Asian–accented ATC dataset of approximately 11 hours and fine‑tuned Whisper‑Small and Whisper‑Large‑v3 Turbo models for accent‑specific adaptation. This training reduced transcription errors compared to baseline Whisper‑ATC models trained on Western accents, achieving a WER of 9.82\% on SEA‑accented speech. Their findings underscore the importance of accent‑aware fine‑tuning in safety‑critical ASR applications.

Recent advancements in ASR have shifted from specialized acoustic models to integrated Speech LLMs. Qwen2.5‑Omni‑7B~\cite{xu2025qwen2} exemplifies this paradigm, coupling a Whisper‑based audio encoder with a 7B‑parameter language model in a “Thinker–Talker” architecture. Unlike traditional encoder–decoder systems focused on phonetic decoding, Speech LLMs leverage semantic reasoning to resolve ambiguities in audio signals. This capability is particularly valuable in ATC, where accurate interpretation of operational phraseology is as essential as clean acoustic transcription.

Finally, Pilmannová~\cite{pilmannova2024incorrect} compared human performance against an automated system for readback error detection, showing that while humans achieved 95.6\% accuracy, the system reached only 59.9\%. Helmke~\cite{helmke2022readback} evaluated both rule‑based and AI‑driven approaches on a proprietary dataset, reporting a detection rate above 80\%. Although these studies laid the groundwork for automated readback monitoring, they relied on cascaded, multi‑stage architectures that feed conventional ASR outputs into separate downstream rule engines or text‑based classifiers (e.g., T5).

To our knowledge, no prior work has applied Speech LLMs directly to ATC tasks. Existing studies rely on cascaded ASR pipelines, multilingual adaptation, or accent‑specific fine‑tuning, but not integrated Speech LLM frameworks. Building on this progression, we design a unified Speech LLM system that jointly performs transcription, role attribution, and structured readback monitoring under realistic aeronautical conditions.

\section{Methodology}

The methodology adopted in this study was designed to capture the complexity of air traffic communication and prepare it for model training and evaluation. Our approach centers on three components: (i) utilization of publicly available ATC datasets to ensure reproducibility, (ii) definition of structured formats that guide supervised fine‑tuning, and (iii) augmentation strategies that broaden conversational coverage. The following subsections detail dataset utilization, format design, and augmentation procedures.

\subsection{Corpus Integration and Metadata Integration}

We employed three publicly available ATC corpora. 

The \textbf{ATCOSIM dataset}\cite{hofbauer-etal-2008-atcosim} comprises simulated controller speech in English, recorded from 10 non‑native speakers. All utterances are produced in the controller role, and the corpus includes explicit disfluency annotations (e.g., the filler marker ``ah'').

The \textbf{UWB‑ATCC dataset}\cite{vsmidl2019air} consists of real‑world Czech‑accented English recordings. Each audio clip is annotated with speaker roles (pilot, controller, or both). We restricted usage to clips labeled as either pilot or controller to support utterance‑level role tagging. We focused our reporting of WER and methodological comparisons on this dataset, ensuring that results reflect realistic operational conditions. For reference, Zuluaga et al.~\cite{zuluaga2022how} reported a WER of 17.56\% on the full dataset.

To further enrich metadata, we incorporated the publicly released \textbf{ATCO2-test-set-1h}~\cite{zuluaga2022atco2}, which comprises approximately 1 hour of audio and transcription data annotated with callsigns, commands, and values. We did not use the ATCO2 audio itself, as its accent distribution differs from UWB‑ATCC. Instead, this supplementary resource enabled us to leverage structured metadata during model training for downstream tasks.

\subsection{Role-based Instruction Tuning}

Beyond conventional audio-text pairs for ASR training, LLMs benefit from structured prompts that explicitly encode task context. Recent work demonstrates the effectiveness of role-based prompting, where assigning domain-specific roles improves adaptation and instruction-following \cite{wang2024rolellm}. Motivated by this insight, our ATC dataset design incorporates role annotations to guide supervised fine‑tuning. Listing~\ref{lst:role_annotation} illustrates the role specification used in the system prompt.

\begin{lstlisting}[caption={LLM role annotation\vspace{0.3em}}, label={lst:role_annotation}]
{
  "system": "You are an expert in Aviation Communications with the experience of a senior Air Traffic Controller. You specialize in ICAO standard phraseology and are tasked with accurately transcribing and analyzing flight data from both controllers and pilots."
}
\end{lstlisting}

We developed two complementary dataset formats. The first pairs each audio file with a JSON entry containing an instruction, input, output, and audio path, enabling multimodal alignment between speech and textual directives. The second consists of text‑only transcripts enriched with ATC tags for callsign, command, and value. Both formats include explicit speaker role annotations (Pilot and ATCO), ensuring consistent role attribution across audio and text. This design enables the model to learn both acoustic grounding and structured communication patterns, improving its ability to handle realistic ATC exchanges and adapt to operational scenarios, which in turn supports downstream tasks such as readback detection. Listings~\ref{lst:json_format} and~\ref{lst:tag_format} illustrate the two formats.

\begin{lstlisting}[caption={Supervised fine‑tuning format with audio reference\vspace{0.3em}}, label={lst:json_format}]
{
  "system": "Refer to Listing 1 for role annotation",
  "instruction": "Transcribe the audio and identify roles.",  
  "input": "<audio>",
  "output": ATCO: sky shuttle one one four zero now descend flight level three three zero",
  "audios": ["/audios/train_0_uwb-atcc_ACCU-0agmXf_000000_000395_AT.wav"]
}
\end{lstlisting}

\begin{lstlisting}[caption={Supervised fine‑tuning format with attribute tags}, label={lst:tag_format}]
{
  "system": "Refer to Listing 1 for role annotation",
   "instruction": "Analyze the ATC transcription. Add back the tags for role, callsign, command, and value",
  "input": "oscar kilo papa mike bravo descend flight level one hundred",
  "output": "[ATCO] [#callsign]Oscar Kilo Papa Mike Bravo[/#callsign][#command]descend[/#command] [#value]flight level one hundred[/#value]",
  "audios": []
}
\end{lstlisting}

\subsection{Data Augmentation}

We applied a simple yet effective augmentation strategy: two audio segments were randomly concatenated to form synthetic conversations. This augmentation forces the model to learn role assignment in multi‑turn exchanges, rather than single isolated utterances. By exposing the model to mixed pilot–controller dialogues, the training data better reflects realistic ATC communication flow and improves robustness in role labeling. Similar concatenation-based augmentation strategies have demonstrated significant improvements in ASR robustness across diverse domains \cite{nguyen2024synthetic}; our adaptation further validates their utility within the highly structured context of ATC communication.


\section{Experiments and Results}

\subsection{Datasets and Models}
For \textbf{ATCOSIM}, we adopted the community split provided on Hugging Face (\texttt{Jzuluaga/atcosim\_corpus}). The total audio length is 10.45 hours. To examine the role of disfluency annotation in ATC‑domain ASR, we analyzed filler labeling: 565 out of 7,646 training utterances contain the marker ``ah,'' highlighting its prevalence and potential impact on recognition accuracy.  

For \textbf{UWB‑ATCC}, we followed the community split available on Hugging Face (\texttt{Jzuluaga/uwb\_atcc}). During evaluation, we excluded short utterances of fewer than four words, as such ultra‑short utterances often lack semantic content and can distort WER estimates. After processing, the corpus comprises approximately 11.5 hours of real ATC recordings.  

For the \textbf{ATCO2-test-set-1h}, a 10\% subset was reserved as a held‑out test set to evaluate model performance on attribute tagging.



We evaluate two representative architectures:

\begin{itemize}
    \item \textbf{Speech LLM (Qwen2.5‑Omni‑7B~\cite{xu2025qwen2})}: a large language model capable of directly processing audio features while leveraging linguistic priors for contextual reasoning. For this model, we applied LoRA adaptation training to the following modules: \texttt{q\_proj}, \texttt{v\_proj}, \texttt{k\_proj}, \texttt{o\_proj}, \texttt{gate\_proj}, \texttt{up\_proj}, \texttt{down\_proj}, \texttt{audio\_projector}, and \texttt{multi\_modal\_projector}, with a LoRA rank of 64, LoRA $\alpha$ of 128, and LoRA dropout of 0.1. 
    \item \textbf{Transformer ASR (Whisper‑large‑v3~\cite{radford2023robust})}: a conventional encoder–decoder speech‑to‑text transformer trained for robust transcription. We performed full‑model fine‑tuning for this baseline.
\end{itemize}


\subsection{Filler Handling in ATC Speech}
We examined the impact of including the discourse marker ``ah'' during training. The results, summarized in Table~\ref{tab:ah_effect}, show how each architecture responds to the presence or absence of this non‑lexical anchor.

\begin{table}[ht]
\centering
\caption{Impact of including ``ah'' token on ASR performance (WER computed with ``ah'' removed from test set)}
\begin{tabular}{lccc}
\hline
\textbf{Model} & \textbf{Training Labels} & \textbf{WER\%} \\
\hline
Speech LLM            & With ah    & 2.49 \\
Speech LLM            & Without ah & 1.18 \\
Transformer ASR      & With ah    & 3.75  \\
Transformer ASR      & Without ah & 3.12  \\
\hline
\end{tabular}
\label{tab:ah_effect}
\end{table}

Our findings extend the observations of \cite{mihajlik2024disfluency} by demonstrating that the effect of fillers varies across architectures and, in the ATC domain, is consistently detrimental. For Transformer‑based ASR (Whisper‑large‑v3), omitting the filler ``ah'' lowers WER (3.12\% vs. 3.75\%), while for the Speech LLM the reduction is even more substantial (1.18\% vs. 2.49\%). Consistent with prior work, fillers degrade recognition accuracy, though the magnitude of this effect depends on both application and model design. Moreover, transcribing fillers requires considerable human effort during annotation, yet the WER results show that this effort yields no tangible benefit.

Building on these findings, we adopt the Speech LLM as the primary model for subsequent experiments, and leveraging its lower WER and reasoning capability for higher‑level tasks such as role assignment and attribute tagging. In later experiments with the ATCOSIM dataset, we omit the filler “ah,” as fillers degrade recognition accuracy. 

\subsection{Impact of Augmentation and Metadata-Enriched Training}

We trained the Speech LLM using three extensions to the Baseline (UWB‑ATCC dataset): (i) the ATCOSIM dataset, which expands ATC‑related training material, (ii) concatenation‑based data augmentation that generates synthetic multi‑turn dialogues, and (iii) a metadata‑enriched transcription dataset pairing transcripts with structured tags (callsign, command, value) and explicit speaker roles. The combined results are summarized in Table~\ref{tab:aug_metadata}.

Training with the ATCOSIM dataset provided a modest reduction in WER, though role accuracy declined slightly, reflecting the influence of simulated controller‑only speech. This addition nonetheless expanded ATC‑related coverage and enriched the training corpus.

The concatenation‑based augmentation strategy yielded consistent improvements: WER decreased and role accuracy slightly increased. By exposing the model to synthetic multi‑turn dialogues, augmentation enhanced its ability to track pilot–controller exchanges, contributing more to discourse‑level robustness than to lexical recognition.

Further gains were achieved by incorporating the metadata‑enriched dataset, which supplies clear cues for segmentation and labeling. This format enables the model to better capture operationally critical information and further reduce WER.

Taken together, these results highlight a complementary effect: Speech LLMs establish strong baseline performance, ATCOSIM adds coverage of controller speech, augmentation improves both transcription accuracy and role assignment, and metadata‑enriched training delivers incremental gains in WER while reinforcing the extraction of operationally critical information.

\begin{table}[ht]
\centering
\caption{Impact of augmentation and metadata‑enriched training on ASR performance}
\begin{tabular}{lccc}
\hline
\textbf{Training Labels} & \textbf{WER\%} & \textbf{Role Accuracy\%} \\
\hline
Baseline (UWB‑ATCC dataset)        & 18.13 & 96.37 \\
+ ATCOSIM dataset                  & 17.79 & 95.14 \\
+ Concatenation‑based augmentation & 16.96 & 96.93 \\
+ Metadata‑enriched transcripts    & 16.70 & 96.60 \\

\hline
\end{tabular}
\label{tab:aug_metadata}
\end{table}



\subsection{Readback Detection}

After training, our model is able to transcribe ATC domain audio with role tags and extract crucial information such as callsigns, commands, and command values. In aviation safety, one critical aspect is the detection of readback. In standard procedure, when a pilot or controller issues a command or request, the receiving party must repeat the key elements of the message. The initiator then listens and confirms that the message has been conveyed correctly.

To evaluate the model's proficiency in extracting these critical attributes, we conducted an empirical assessment using the ATCO2 dataset. The 10\% held-out test set comprises 56 samples, with 21 instances achieving perfectly correct extraction across all callsign, command, and value attributes simultaneously. Precision, recall and F1-score were calculated and are summarized in Table~\ref{tab:model_performance}. While the model demonstrates a high F1-score for callsigns (0.955), its performance drops noticeably for commands (0.667) and values (0.674). This stark drop from the training set F1-scores (0.997 for callsigns, 0.976 for commands, and 0.990 for values) underscores the challenges the model faces when encountering unseen cases.

We suggest that these evaluation discrepancies arise primarily from systemic ambiguities and inconsistencies within the ground‑truth annotations, rather than reflecting intrinsic limitations of the model itself. For instance, the gold standard labels ``at one four knots'' as a value, whereas the model predicts ``one four knots,'' resulting in a technical misclassification despite their complete semantic equivalence. In another case, the ground truth splits a sequence into ``report passing'' [command] and ``one three thousand feet'' [value], while the model yields ``report'' [command] and ``passing one three thousand feet'' [value]. Furthermore, explicit annotation quality issues are present in the dataset. For example, the ground truth labels a specific wind-check segment as \texttt{wind check for [\#value]runway one four three[/\#value] [\#value]four degrees[/\#value] [\#value]two knots[/\#value]}. However, according to standard international aviation phraseology, runway designators are strictly restricted to two digits, and wind directions must be transmitted as three-digit headings. The gold standard's inclusion of ``three'' into the runway designator creates an impossible runway (143) and leaves the subsequent wind direction ``four degrees'' highly ambiguous. The model successfully adheres to domain logic by correctly separating ``one four'' as the runway, yet it is mathematically penalized by the strict token matching of the evaluation script. These alignment mismatches indicate that data annotation noise and subjective parsing boundaries artificially suppress the measured performance metrics.

\begin{table}[htbp]
\centering
\caption{Classification Performance Metrics by Attribute Category}
\label{tab:model_performance}
\begin{tabular}{lcccccc}
\hline
\textbf{Category} & \textbf{Precision} & \textbf{Recall} & \textbf{F1-Score}\\
\hline
CALLSIGN & 0.938 & 0.974 & 0.955   \\
COMMAND  & 0.632 & 0.706 & 0.667 \\
VALUE    & 0.664 & 0.684 & 0.674 \\
\hline
\end{tabular}
\end{table}

\begin{figure}[tbp]
    \centering
    \includegraphics[width=0.5\textwidth]{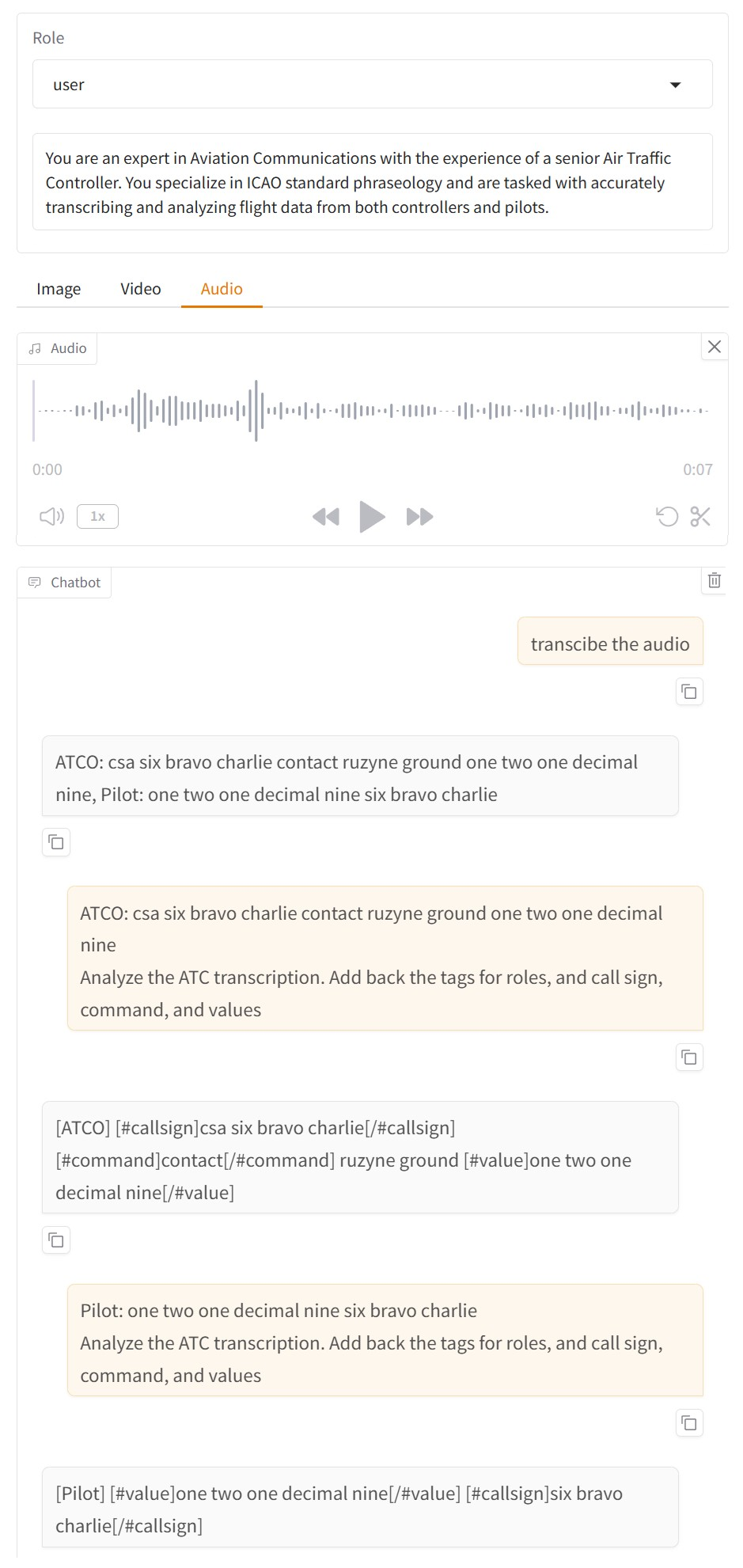}
    \caption{Example of real-time readback detection enabled by our model.}
    \label{fig:ourtts}
\end{figure}

Despite these annotation‑induced metric penalties, the model demonstrates robust semantic reasoning and high utility in operational contexts. To illustrate this, we selected the audio clip \textit{10896\_uwb‑atcc\_TWR‑4nID1r\_000098\_000766\_PIAT.wav} from the UWB‑ATCC dataset. This example highlights the system’s capacity to support real‑time readback verification. As shown in Figure~\ref{fig:ourtts}, the conversational audio sample was processed by the trained model. The first instruction was to transcribe the audio, and based on this output, follow‑up instructions were applied to insert the appropriate attribute tags.

The ground‑truth transcription is:

\noindent\textit{ATCO: csa six bravo charlie contact ruzyne ground one two one decimal nine. Pilot: nine one two one decimal nine six bravo charlie.}

Our model produced a highly accurate transcription. To post‑process the output, we applied simple rules to segment the text by speaker roles and then prompted the model to insert the appropriate tags. The extracted elements for callsign, command, and numerical values are summarized in Table~\ref{tab:readback_case1}.

\begin{table}[htbp]
\centering
\caption{Case study 1: Comparison of ATCO and Pilot utterances with extracted tags (no readback error)}
\label{tab:readback_case1}
\begin{tabular}{l|c|c}
\hline
\textbf{Element} & \textbf{ATCO} & \textbf{Pilot} \\
\hline
Callsign & csa six bravo charlie & six bravo charlie \\
Command  & contact               & -- \\
Value    & one two one decimal nine & one two one decimal nine \\
\hline
\end{tabular}
\end{table}

By comparing the extracted elements, we confirm that both sides reference the same callsign, “csa six bravo charlie,” although the pilot omits the airline prefix “csa.” The controller issues the command “contact,” and the pilot correctly repeats the crucial numerical value “one two one decimal nine.” Thus, no readback error is present in this case, as the essential numerical information was acknowledged and confirmed.

We further evaluated an audio sample from the ATCO2 project, which illustrates a readback error detected by our framework. The results are presented in Table~\ref{tab:readback_case2}. 

The ground‑truth transcription is:

\noindent\textit{ATCO: easy six nine four one descend flight level one one zero. Pilot: descending flight level one three zero easy six nine four one.}




Our model produced a highly accurate transcription. After structured extraction, we note that the command is “descend.” The command value is 110, while the pilot repeated it as 130. Our approach therefore successfully detects this readback error.

\begin{table}[htbp]
\centering
\caption{Case study 2: Comparison of ATCO and Pilot utterances with extracted tags (readback error detected)}
\label{tab:readback_case2}
\begin{tabular}{l|c|c}
\hline
\textbf{Element} & \textbf{ATCO} & \textbf{Pilot} \\
\hline
Callsign & easy six nine four one & easy six nine four one \\
Command   & descend                     & descending \\
Value     & flight level one one zero   & flight level one three zero \\
\hline
\end{tabular}
\end{table}

These case studies demonstrate the potential of automated readback detection to enhance ATC safety monitoring and support operational reliability.

To put the system into operational use, audio streams can be fed directly into the model. Standard voice activity detection (VAD) is applied to mark the end of each speech segment, after which transcription is generated. The output is then checked for role tags, and for each tagged utterance, key information such as callsigns, commands, and values is extracted and stored in a database. With a benchmarked latency of approximately 0.8 seconds per LLM call (and around 2.4 seconds for sequential three‑call workflows), the system delivers low‑latency processing suitable for real‑time deployment. During insertion, recent entries associated with the same callsign and command can be retrieved, enabling the system to automatically compare controller and pilot values and identify potential inconsistencies in real time, thereby advancing proactive air traffic safety through transcription, attribute tagging, and readback detection.

\section{Conclusion}

In this work, we introduced a domain‑specific approach for transcribing and structuring ATC communications using a Speech LLM. The model achieves accurate transcription with explicit role tagging and extracts key operational elements such as callsigns, commands, and values. Through a case study on readback detection, we demonstrated how structured outputs enable systematic comparison between controller and pilot utterances, facilitating the identification of potential readback errors, a critical factor in aviation safety. By integrating transcription, segmentation, and structured tagging into a unified framework, our Speech LLM approach establishes a foundation for real‑time readback detection and paves the way for operational deployment in ATC monitoring, where automated systems can proactively support safety oversight.


\section*{Acknowledgment}

This work was supported by a grant from 
the Innovation and Technology Fund of the Hong Kong SAR, China (Project No: ITT/028/24LP).

\bibliographystyle{IEEEtran} 
\bibliography{references}    

@article{khalil2023automatic,
  title={An automatic speaker clustering pipeline for the air traffic communication domain},
  author={Khalil, Driss and Prasad, Amrutha and Motlicek, Petr and Zuluaga-Gomez, Juan and Nigmatulina, Iuliia and Madikeri, Srikanth and Schuepbach, Christof},
  journal={Aerospace},
  volume={10},
  number={10},
  pages={876},
  year={2023},
  publisher={MDPI}
}

@article{vsmidl2019air,
  title={Air traffic control communication (ATCC) speech corpora and their use for ASR and TTS development},
  author={{\v{S}}m{\'\i}dl, Lubo{\v{s}} and {\v{S}}vec, Jan and Tihelka, Daniel and Matou{\v{s}}ek, Jind{\v{r}}ich and Romportl, Jan and Ircing, Pavel},
  journal={Language Resources and Evaluation},
  pages={449--464},
  year={2019},
  publisher={JSTOR}
}

@article{zuluaga2022how,
    title={How Does Pre-trained Wav2Vec2. 0 Perform on Domain Shifted ASR? An Extensive Benchmark on Air Traffic Control Communications},
    author={Zuluaga-Gomez, Juan and Prasad, Amrutha and Nigmatulina, Iuliia and Sarfjoo, Saeed and others},
    journal={IEEE Spoken Language Technology Workshop (SLT), Doha, Qatar},
    year={2022}
  }

@inproceedings{hofbauer-etal-2008-atcosim,
    title = "The {ATCOSIM} Corpus of Non-Prompted Clean Air Traffic Control Speech",
    author = "Hofbauer, Konrad  and
      Petrik, Stefan  and
      Hering, Horst",
    editor = "Calzolari, Nicoletta  and
      Choukri, Khalid  and
      Maegaard, Bente  and
      Mariani, Joseph  and
      Odijk, Jan  and
      Piperidis, Stelios  and
      Tapias, Daniel",
    booktitle = "Proceedings of the Sixth International Conference on Language Resources and Evaluation ({LREC}'08)",
    month = may,
    year = "2008",
    address = "Marrakech, Morocco",
    publisher = "European Language Resources Association (ELRA)",
    url = "https://aclanthology.org/L08-1507/"
}

@article{blatt2024joint,
  title={Joint vs sequential speaker-role detection and automatic speech recognition for air-traffic control},
  author={Blatt, Alexander and Krishnan, Aravind and Klakow, Dietrich},
  journal={arXiv preprint arXiv:2406.13842},
  year={2024}
}

@article{baevski2020wav2vec,
  title={wav2vec 2.0: A framework for self-supervised learning of speech representations},
  author={Baevski, Alexei and Zhou, Yuhao and Mohamed, Abdelrahman and Auli, Michael},
  journal={Advances in neural information processing systems},
  volume={33},
  pages={12449--12460},
  year={2020}
}

@article{xu2025joint,
  title={Joint ASR and Speaker Role Tagging with Serialized Output Training},
  author={Xu, Anfeng and Feng, Tiantian and Narayanan, Shrikanth},
  journal={arXiv preprint arXiv:2506.10349},
  year={2025}
}

@inproceedings{radford2023robust,
  title={Robust speech recognition via large-scale weak supervision},
  author={Radford, Alec and Kim, Jong Wook and Xu, Tao and Brockman, Greg and McLeavey, Christine and Sutskever, Ilya},
  booktitle={International conference on machine learning},
  pages={28492--28518},
  year={2023},
  organization={PMLR}
}

@article{xu2025qwen2,
  title={Qwen2. 5-Omni Technical Report},
  author={Xu, Jin and Guo, Zhifang and He, Jinzheng and Hu, Hangrui and He, Ting and Bai, Shuai and Chen, Keqin and Wang, Jialin and Fan, Yang and Dang, Kai and others},
  journal={arXiv preprint arXiv:2503.20215},
  year={2025}
}

@inproceedings{wee2025adapting,
  title={Adapting Automatic Speech Recognition for Accented Air Traffic Control Communications},
  author={Wee, Marcus Yu Zhe and Wong, Justin Juin Hng and Lim, Lynus and Tan, Joe Yu Wei and Gupta, Prannaya and Lim, Dillion and Tew, En Hao and Han, Aloysius Keng Siew and Lim, Yong Zhi},
  booktitle={2025 International Conference on Military Communication and Information Systems (ICMCIS)},
  pages={1--10},
  year={2025},
  organization={IEEE}
}

@inproceedings{devlin2019bert,
  title={Bert: Pre-training of deep bidirectional transformers for language understanding},
  author={Devlin, Jacob and Chang, Ming-Wei and Lee, Kenton and Toutanova, Kristina},
  booktitle={Proceedings of the 2019 conference of the North American chapter of the association for computational linguistics: human language technologies, volume 1 (long and short papers)},
  pages={4171--4186},
  year={2019}
}

@inproceedings{mihajlik2024disfluency,
  title={On Disfluency and Non-lexical Sound Labeling for End-to-end Automatic Speech Recognition.},
  author={Mihajlik, Peter and Meng, Yan and Kadar, Mate S and Linke, Julian and Schuppler, Barbara and M{\'a}dy, Katalin},
  booktitle={Interspeech},
  year={2024}
}

@article{zuluaga2022atco2,
  title={Atco2 corpus: A large-scale dataset for research on automatic speech recognition and natural language understanding of air traffic control communications},
  author={Zuluaga-Gomez, Juan and Vesel{\`y}, Karel and Sz{\"o}ke, Igor and Blatt, Alexander and Motlicek, Petr and Kocour, Martin and Rigault, Mickael and Choukri, Khalid and Prasad, Amrutha and Sarfjoo, Seyyed Saeed and others},
  journal={arXiv preprint arXiv:2211.04054},
  year={2022}
}

@inproceedings{wang2024rolellm,
  title={Rolellm: Benchmarking, eliciting, and enhancing role-playing abilities of large language models},
  author={Wang, Noah and Peng, Zy and Que, Haoran and Liu, Jiaheng and Zhou, Wangchunshu and Wu, Yuhan and Guo, Hongcheng and Gan, Ruitong and Ni, Zehao and Yang, Jian and others},
  booktitle={Findings of the Association for Computational Linguistics: ACL 2024},
  pages={14743--14777},
  year={2024}
}

@inproceedings{nguyen2024synthetic,
  title={Synthetic conversations improve multi-talker ASR},
  author={Nguyen, Thai-Binh and Waibel, Alexander},
  booktitle={ICASSP 2024-2024 IEEE International Conference on Acoustics, Speech and Signal Processing (ICASSP)},
  pages={10461--10465},
  year={2024},
  organization={IEEE}
}

@article{pilmannova2024incorrect,
  title={Incorrect Read-back Detection in Air Traffic Control},
  author={Pilmannov{\'a}, Ter{\'e}zia and Kalanin, Adam and Schmidt, Stanley},
  journal={Transportation Research Procedia},
  volume={81},
  pages={256--265},
  year={2024},
  publisher={Elsevier}
}

@article{helmke2022readback,
  title={Readback error detection by automatic speech recognition and understanding},
  author={Helmke, Hartmut and Ondrej, K and Shetty, Shruthi and Aril{\'\i}usson, H{\"o}r{\dh}ur and Simiganoschi, T and Kleinert, Matthias and Ohneiser, Oliver and Her, H and Zuluaga-Gomez, Juan and Smrz, Pavel},
  journal={Proceedings of the 12th SESAR Innovation Days, Budapest, Hungary},
  pages={5--8},
  year={2022}
}

\end{document}